%% file: egpaper_for_review.tex
\documentclass[10pt,twocolumn,letterpaper]{article}

\usepackage{cvpr}
\usepackage{times}
\usepackage{epsfig}
\usepackage{graphicx}
\usepackage{amsmath}
\usepackage{booktabs}
\usepackage{amssymb}
\usepackage[misc]{ifsym}
\usepackage[bottom]{footmisc}
\usepackage[linesnumbered,ruled]{algorithm2e}
\usepackage{float}

\usepackage[x11names,dvipsnames,table]{xcolor} 
\usepackage{multirow}
\usepackage{array}
\newcolumntype{L}[1]{>{\raggedright\let\newline\\\arraybackslash\hspace{0pt}}m{#1}}
\newcolumntype{C}[1]{>{\centering\let\newline\\\arraybackslash\hspace{0pt}}m{#1}}
\newcolumntype{R}[1]{>{\raggedleft\let\newline\\\arraybackslash\hspace{0pt}}m{#1}}

\usepackage[pagebackref=true,breaklinks=true,letterpaper=true,colorlinks,bookmarks=false]{hyperref}

\cvprfinalcopy 

\def\cvprPaperID{1431} 

\ifcvprfinal\fi
\begin{document}

\title{Adversarial Attacks Beyond the Image Space}

\author{
Xiaohui Zeng$^1$, Chenxi Liu$^2$$^{(\textrm{\Letter})}$, Yu-Siang Wang$^3$, Weichao Qiu$^2$, \\ Lingxi Xie$^{2,4}$, Yu-Wing Tai$^5$, Chi-Keung Tang$^6$, Alan L. Yuille$^2$\\
$^1$University of Toronto\quad$^2$The Johns Hopkins University\quad$^3$National Taiwan University\\
$^4$Huawei Noah's Ark Lab\quad$^5$Tencent YouTu\quad$^6$Hong Kong University of Science and Technology\\
{\tt\small xiaohui@cs.toronto.edu}\qquad{\tt\small cxliu@jhu.edu}\qquad{\tt\small b03202047@ntu.edu.tw}\\
{\tt\small\{qiuwch, 198808xc, yuwing, alan.l.yuille\}@gmail.com}\qquad{\tt\small cktang@cs.ust.hk}
}

\maketitle

\begin{abstract}
\vspace{-0.15cm}
Generating adversarial examples is an intriguing problem and an important way of understanding the working mechanism of deep neural networks. Most existing approaches generated perturbations in the image space, i.e., each pixel can be modified independently. However, in this paper we pay special attention to the subset of adversarial examples that correspond to meaningful changes in 3D physical properties (like rotation and translation, illumination condition, etc.). These adversaries arguably pose a more serious concern, as they demonstrate the possibility of causing neural network failure by easy perturbations of real-world 3D objects and scenes.

In the contexts of object classification and visual question answering, we augment state-of-the-art deep neural networks that receive 2D input images with a rendering module (either differentiable or not) in front, so that a 3D scene (in the physical space) is rendered into a 2D image (in the image space), and then mapped to a prediction (in the output space). The adversarial perturbations can now go beyond the image space, and have clear meanings in the 3D physical world. Though image-space adversaries can be interpreted as per-pixel albedo change, we verify that they cannot be well explained along these physically meaningful dimensions, which often have a non-local effect. But it is still possible to successfully attack beyond the image space on the physical space, though this is more difficult than image-space attacks, reflected in lower success rates and heavier perturbations required.
\end{abstract}

\vspace{-0.2cm}
\section{Introduction}
\label{Introduction}

\begin{figure}[t]
\begin{center}
\includegraphics[width=\linewidth]{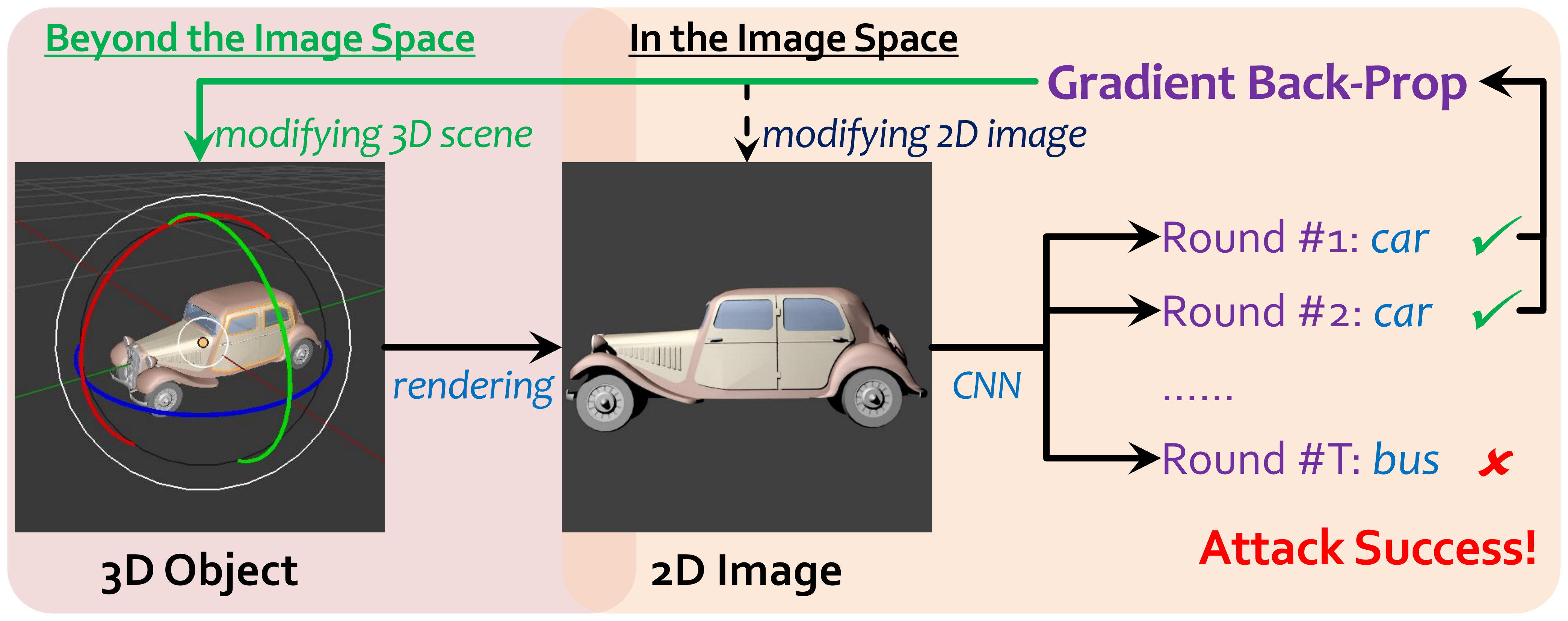}
\end{center}
\caption{
The vast majority of existing works on adversarial attacks focus on modifying pixel values in 2D images to cause wrong CNN predictions. In our work, we consider the more complete vision pipeline, where 2D images are in fact projections of the underlying 3D scene. This suggests that adversarial attacks can go \emph{beyond} the image space, and directly change physically meaningful properties that define the 3D scene. We suspect that these adversarial examples are more physically plausible and thus pose more serious security concerns.
}
\label{Fig:Framework}
\vspace{-0.1cm}
\end{figure}

\begin{figure*}[t]
\begin{center}
\includegraphics[width=0.875\linewidth]{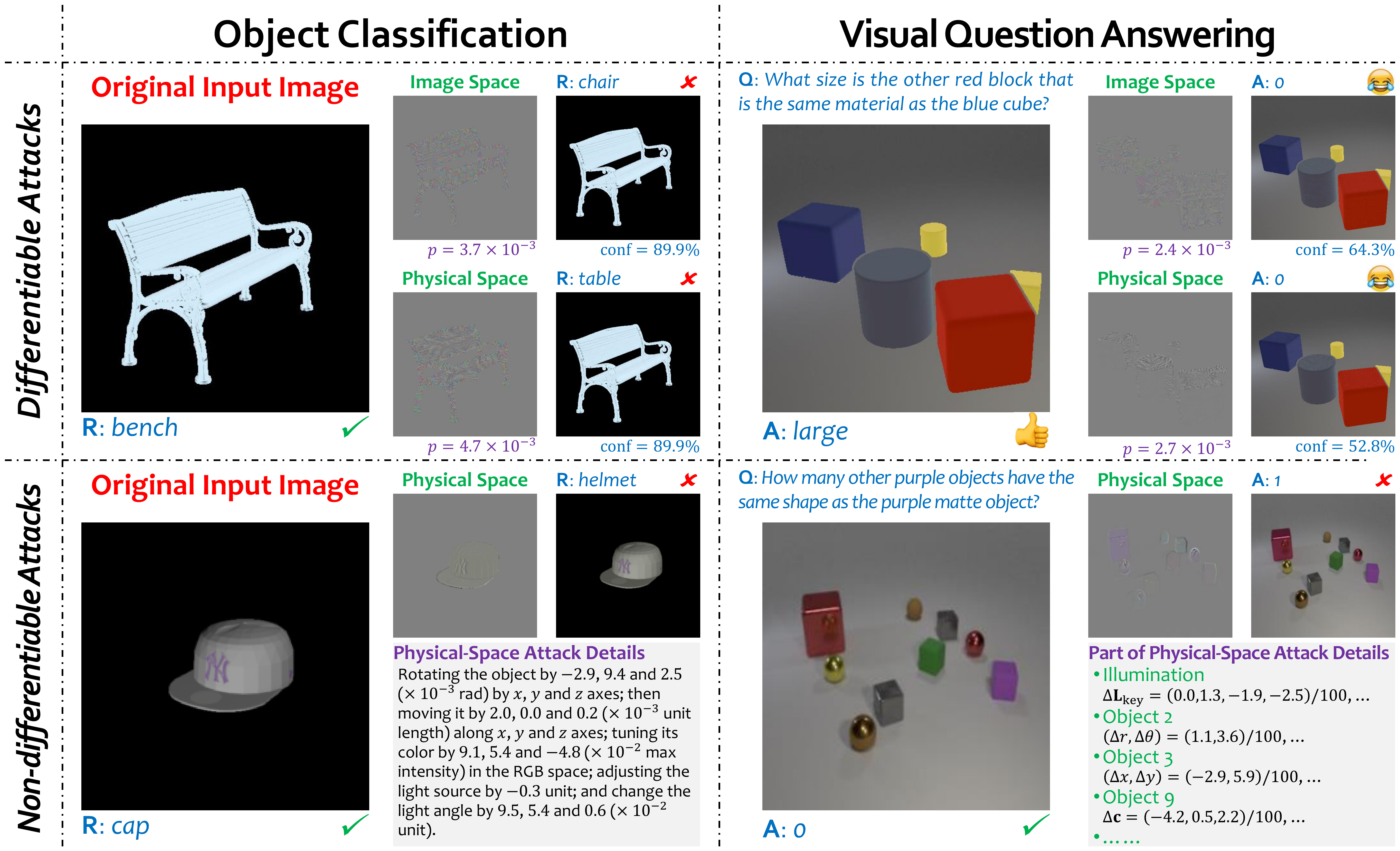}
\end{center}
\caption{
Adversarial examples for 3D object classification and visual question answering, under either a differentiable or a non-differentiable renderer. The top row shows that while it is of course possible to produce adversarial examples by attacking the image space, it is also possible to successfully attack on the physical space by changing factors such as surface normal, material, lighting condition (see Section~\ref{Approach:Network}). The bottom row demonstrates the same using a more realistic non-differentiable renderer, with descriptions of how to carry out the attack. $p$ and $\mathrm{conf}$ are the perceptibility (see Section~\ref{Approach:Attacks}) and the confidence (post-softmax output) on the predicted class.
}
\label{Fig:Motivation}
\end{figure*}

Recent years have witnessed a rapid development in the area of deep learning, in which deep neural networks have been applied to a wide range of computer vision tasks, such as image classification~\cite{Krizhevsky_2012_ImageNet}\cite{He_2016_Deep}, object detection~\cite{Ren_2017_Faster}, semantic segmentation~\cite{Shelhamer_2017_Fully}\cite{Chen_2017_DeepLab}, visual question answering~\cite{Antol_2015_VQA}\cite{Johnson_2017_CLEVR}, {\em etc}.
Despite the great success of deep learning, there still lacks an effective method to understand the working mechanism of deep neural networks. An interesting effort is to generate so-called {\em adversarial perturbations}. They are visually imperceptible noise~\cite{Goodfellow_2015_Explaining} which, after being added to an input image, changes the prediction results completely, sometimes ridiculously. These examples can be constructed in a wide range of vision problems, including image classification~\cite{Nguyen_2015_Deep}, object detection and semantic segmentation~\cite{Xie_2017_Adversarial}. Researchers believed that the existence of adversaries implies unknown properties in the feature space~\cite{Szegedy_2014_Intriguing}.

Our work is motivated by the fact that conventional 2D adversaries were often generated by modifying each image pixel individually. We instead consider perturbations of the 3D scene that are often non-local and correspond to physical properties of the object. We notice that previous work found adversarial examples ``in the physical world'' by taking photos on the printed perturbed images~\cite{Kurakin_2017_Adversarial1}. But our work is different and more essential, as we are attacking the intrinsic parameters that define the 3D scene/object, whereas~\cite{Kurakin_2017_Adversarial1} is still limited to attacking 2D image pixels. For this respect, we plug 3D rendering as a network module into the state-of-the-art neural networks for object classification and visual question answering. In this way, we build a mapping function from the {\em physical space} (a set of physical parameters, including surface normals, illumination and material), via the {\em image space} (a rendered 2D image), to the {\em output space} (the object class or the answer to a question). See Figure~\ref{Fig:Framework} which illustrates this framework. 

The per-pixel image-space attack can be explained in terms of per-pixel changes of albedo, but it is highly unlikely that these individual perturbations happen to correspond to, e.g., a simple rotation of the object in 3D. 
Using our pipeline with rendering, we indeed found it almost impossible to approximate the 2D image adversaries using the 3D physically meaningful perturbations. At the same time, this suggests a natural mechanism for defending adversaries -- finding an approximate solution in the physical space and re-rendering will make most image-space adversaries fail. This analysis-by-synthesis  process offers new direction in dealing with adversarial examples and occlusion cases.

Our paper mainly tries to answer the following question: 
{\bf can neural networks still be fooled if we do not perturb 2D image pixels, but instead perturb 3D physical properties?} This is about directly generating perturbations in the physical space ({\em i.e.}, modifying basic physical parameters) that cause the neural network predictions to fail. Specifically, we compute the difference between the current output and the desired output, and use gradient descent to update parameters in the physical space ({\em i.e.}, \emph{beyond} the image space, which contains physical parameters such as surface normals and illumination conditions). This attack is implemented by either iterative Fast Gradient Sign Method (FGSM)~\cite{Goodfellow_2015_Explaining} (for differentiable rendering) or the Zeroth-Order Optimization approach~\cite{Chen_2017_ZOO} (for non-differentiable rendering). We constrain the change in the image intensities to guarantee the perturbations to be visually imperceptible. Our major finding is that attacking the physical space is more difficult than attacking the image space. Although it is possible to find adversaries in this way (see Figure~\ref{Fig:Motivation} for a few of these examples), the success rate is lower and the perceptibility of perturbations becomes much larger than required in the image space. This is expected, as the rendering process couples changes in pixel values, {\em i.e.}, modifying one physical parameter ({\em e.g.}, illumination) may cause many pixels to be changed at the same time.

\section{Related Work}
\label{RelatedWork}

Deep learning is the state-of-the-art machine learning technique to learn visual representations from labeled data. Yet despite the success of deep learning, it remains challenging to explain what is learned by these complicated models. One of the most interesting evidence is {\em adversaries}~\cite{Goodfellow_2015_Explaining}: small noise that is (i) imperceptible to humans, and (ii) able to cause deep neural networks make wrong predictions after being added to the input image. Early studies were mainly focused on image classification~\cite{Nguyen_2015_Deep}\cite{Moosavi_2015_DeepFool}. But soon, researchers were able to attack deep networks for detection and segmentation~\cite{Xie_2017_Adversarial}, and also visual question answering \cite{Xu_2017_Can}. Efforts were also made in finding universal perturbations which can transfer across images~\cite{Moosavi_2017_Universal}, as well as adversarial examples in the physical world produced by taking photos on the printed perturbed images~\cite{Kurakin_2017_Adversarial1}.

Attacking a known network (both network architecture and weights are given, {\em a.k.a}, a white box) started with setting a goal. There were generally two types of goals. The first one (a non-targeted attack) aimed at reducing the probability of the true class~\cite{Nguyen_2015_Deep}, and the second one (a targeted attack) defined a specific class that the network should predict~\cite{Liu_2017_Delving}. After that, the error between the current and the target predictions was computed, and gradients back-propagated to the image layer. This idea was developed into a set of algorithms, including the Steepest Gradient Descent Method (SGDM)~\cite{Moosavi_2015_DeepFool} and the Fast Gradient Sign Method (FGSM)~\cite{Goodfellow_2015_Explaining}. The difference lies in that SGDM computed accurate gradients, while FGSM merely kept the sign in every dimension. 
The iterative version of these two algorithms were also studied~\cite{Kurakin_2017_Adversarial1}. In comparison, attacking an unknown network ({\em a.k.a.}, a black box) is much more challenging~\cite{Liu_2017_Delving}, and an effective way is to sum up perturbations from a set of white-box attacks~\cite{Xie_2017_Adversarial}.
In opposite, there exist efforts in protecting deep networks from adversarial attacks~\cite{Papernot_2016_Distillation}\cite{Kurakin_2017_Adversarial2}\cite{Tramer_2017_Ensemble}. People also designed algorithms to hack these defenders~\cite{Carlini_2017_Towards} as well as to detect whether adversarial attacks are present~\cite{Metzen_2017_Detecting}. This competition has boosted both attackers and defenders to a higher level~\cite{Athalye_2018_Obfuscated}.

More recently, there is increasing interest in adversarial attacks other than modifying pixel values. \cite{Kurakin_2017_Adversarial1} showed that the adversarial effect still exists if we print the digitally-perturbed 2D image on paper. \cite{Engstrom_2017_Rotation}\cite{Pei_2017_Towards} fooled vision systems by rotating the 2D image or changing its brightness. \cite{Evtimov_2017_Robust}\cite{Athalye_2018_Synthesizing} created real-world 3D objects, either by 3D printing or applying stickers, that consistently cause perception failure. However, these adversaries have high perceptibility and must involve sophisticated change in object appearance. To find adversarial examples in 3D, we use a renderer, either differentiable or non-differentiable, to map a 3D scene to a 2D image and then to the output. In this way it is possible, though challenging, to generate interpretable and physically plausible adversarial perturbations in the 3D scene.

\section{Approach}
\label{Approach}

\subsection{From Physical Parameters to Prediction}
\label{Approach:Network}

As the basis of this work, we extend deep neural networks to receive the physical parameters of a 3D scene, render them into a 2D image, and output prediction, {\em e.g.}, the class of an object, or the answer to a visual question. Note that our research involves 3D to 2D rendering as part of the pipeline, which stands out from previous work which either worked on rendered 2D images~\cite{Su_2015_Multi}\cite{Johnson_2017_Inferring}, or directly processed 3D data without rendering them into 2D images~\cite{Qi_2017_PointNet}\cite{Sfikas_2017_Exploiting}.

We denote the physical space, image space and output space by $\mathcal{X}$, $\mathcal{Y}$ and $\mathcal{Z}$, respectively. Given a 3D scene ${\mathbf{X}}\in{\mathcal{X}}$, the first step is to render it into a 2D image ${\mathbf{Y}}\in{\mathcal{Y}}$, and the second step is to predict the output of $\mathbf{Y}$, denoted by ${\mathbf{Z}}\in{\mathcal{Z}}$. The overall framework is denoted by ${\mathbf{Z}}={\mathbf{f}\!\left[\mathbf{r}\!\left(\mathbf{X}\right);\boldsymbol{\theta}\right]}$, where $\mathbf{r}\!\left(\cdot\right)$ is the renderer, $\mathbf{f}\!\left[\cdot;\boldsymbol{\theta}\right]$ is the target deep network with $\boldsymbol{\theta}$ being parameters.

There are different models for the {\bf 3D rendering function} $\mathbf{r}\!\left(\cdot\right)$. One of them is {\em differentiable}~\cite{Liu_2017_Material}, which considers three sets of physical parameters, {\em i.e.}, surface normals $\mathbf{N}$, illumination $\mathbf{L}$, and material $\mathbf{m}$\footnote{In this model, $\mathbf{N}$ is a $2$-channel image of spatial size $W_\mathrm{N}\times H_\mathrm{N}$, where each pixel is encoded by the azimuth and polar angles of the normal vector at this position; $\mathbf{L}$ is defined by an HDR environment map of dimension $W_\mathrm{L}\times H_\mathrm{L}$, with each pixel storing the intensity of the light coming from this direction (a spherical coordinate system is used); and $\mathbf{m}$ impacts image rendering with a set of bidirectional reflectance distribution functions (BRDFs) which describe the point-wise light reflection for both diffuse and specular surfaces~\cite{Nicodemus_1992_Geometrical}. The material parameters used in this paper come from the directional statistics BRDF model~\cite{Nishino_2009_Directional}, which represents a BRDF as a combination of $D_\mathrm{m}$ distributions with $P_\mathrm{m}$ parameters in each. Mathematically, we have ${\mathbf{N}}\in{\mathbb{R}^{W_\mathrm{N}\times H_\mathrm{N}\times2}}$, ${\mathbf{L}}\in{\mathbb{R}^{W_\mathrm{L}\times H_\mathrm{L}}}$ and ${\mathbf{m}}\in{\mathbb{R}^{D_\mathrm{m}\times P_\mathrm{m}}}$.}. By giving these parameters, we assume that the camera geometries, {\em e.g.}, position, rotation, field-of-view, {\em etc.}, are known beforehand and will remain unchanged in each case. The rendering module is denoted by ${\mathbf{Y}}={\mathbf{r}\!\left(\mathbf{N},\mathbf{L},\mathbf{m}\right)}$. In practice, the rendering process is implemented as a network layer, which is differentiable to input parameters $\mathbf{N}$, $\mathbf{L}$ and $\mathbf{m}$. Another option is to use a {\em non-differentiable} renderer which often provides much higher quality~\cite{Blender_2017_Blender}\cite{McCormac_2017_SceneNet}. In practice we choose an open-source software named Blender~\cite{Blender_2017_Blender}. Not assuming differentiability makes it possible to work on a wider range of parameters, such as color ({\bf C}), translation ({\bf T}), rotation ({\bf R}) and lighting ({\bf L}) considered in this work, in which translation and rotation cannot be implemented by a differentiable renderer\footnote{For 3D object classification, we follow~\cite{Su_2015_Multi} to configure the 3D scene. $\mathbf{L}$ is a $5$-dimensional vector, where the first two dimensions indicate the magnitudes of the environment and point light sources, and the last three the position of the point light source. $\mathbf{C}$, $\mathbf{T}$, $\mathbf{R}$ are all $3$-dimensional properties of the single object. For 3D visual question answering we follow~\cite{Johnson_2017_CLEVR}. $\mathbf{L}$ is a $12$-dimensional vector that represents the energy and position of $3$ point light sources. For every object in the scene, $\mathbf{C}$ is $3$-dimensional, corresponding to RGB; $\mathbf{T}$ is $2$-dimensional which is the object's 2D location on the plane; $\mathbf{R}$ is a scalar rotation angle.}.

We consider two popular {\bf object understanding tasks}, namely, 3D object classification and 3D visual question answering, both of which are straightforward based on the rendered 2D images. Object classification is built upon standard deep networks, and visual question answering, when both the input image $\mathbf{Y}$ and question $\mathbf{q}$ are given, is also a variant of image classification (the goal is to choose the correct answer from a pre-defined set of choices).

In the adversary generation stage, given pre-trained networks, the goal is to attack a model ${\mathbf{Z}}={\mathbf{f}\!\left[\mathbf{r}\!\left(\mathbf{X}\right);\boldsymbol{\theta}\right]}={\mathbf{f}\circ\mathbf{r}\!\left(\mathbf{X};\boldsymbol{\theta}\right)}$. For object classification, $\boldsymbol{\theta}$ is fixed network weights, denoted by $\boldsymbol{\theta}^\mathrm{C}$. For visual question answering, it is weights from an assembled network determined by the question $\mathbf{q}$, denoted by $\boldsymbol{\theta}^\mathrm{V}\!\left(\mathbf{q}\right)$. ${\mathbf{Z}}\in{\left[0,1\right]^K}$ is the output, with $K$ being the number of object classes or choices.

\subsection{Attacks Beyond the Image Space}
\label{Approach:Attacks}

Attacking the physical parameters starts with setting a {\em goal}, which is what we hope the network to predict. This is done by minimizing a loss function $\mathcal{L}\!\left(\mathbf{Z}\right)$, which determines how far the current output is from the desired status. An adversarial attack may either be targeted or non-targeted, and in this work we focus on the non-targeted attack, which specifies a class $c'$ (usually the original true class) as which the image should {\em not} be classified, and the goal is to minimize the $c'$-th dimension of the output $\mathbf{Z}$: ${\mathcal{L}\!\left(\mathbf{Z}\right)}\doteq{\mathcal{L}\!\left(\mathbf{Z};c'\right)}={Z_{c'}}$.

An obvious way to attack the physical space works by expanding the loss function $\mathcal{L}\!\left(\mathbf{Z}\right)$, {\em i.e.}, ${\mathcal{L}\!\left(\mathbf{Z}\right)}={\mathcal{L}\circ\mathbf{f}\circ\mathbf{r}\!\left(\mathbf{X};\boldsymbol{\theta}\right)}$, and minimizing this function with respect to the physical parameters $\mathbf{X}$. The optimization starts with an initial (unperturbed) state ${\mathbf{X}_0}\doteq{\mathbf{X}}$. A total of $T_\mathrm{max}$ iterations are performed. In the $t$-th round, we compute the gradient vectors with respect to $\mathbf{X}_{t-1}$, {\em i.e.}, ${\Delta\mathbf{X}_t}={\nabla_{\mathbf{X}_{t-1}}\mathcal{L}\circ\mathbf{f}\circ\mathbf{r}\!\left(\mathbf{X}_{t-1},\boldsymbol{\theta}\right)}$, and update $\mathbf{X}_{t-1}$ along this direction: ${\mathbf{X}_t}={\mathbf{X}_{t-1}+\eta\cdot\Delta\mathbf{X}_{t-1}}$, where $\eta$ is the {\em learning rate}. This iterative process is terminated if the goal of attacking is achieved or the maximal number of iterations $T_\mathrm{max}$ is reached. The accumulated perturbation over all $T$ iterations is denoted by ${\Delta\mathbf{X}}={\eta\cdot{\sum_{t=1}^T}\Delta\mathbf{X}_t}$.

The way of computing gradients $\Delta\mathbf{X}_t$ depends on whether $\mathbf{r}\!\left(\cdot\right)$ is differentiable. If so, this can be simply back-propagate gradients from the output space to the physical space. We follow the Fast Gradient Sign Method (FGSM)~\cite{Goodfellow_2015_Explaining} to only preserve the sign in each dimension of the gradient vector. Otherwise, we apply zeroth-order optimization. To attack the $d$-th dimension in $\mathbf{X}$, we set a small value $\delta$ and approximate the gradient of $\mathbf{Z}$ by ${\frac{\partial\mathcal{L}\!\left(\mathbf{Z}\right)}{\partial X_d}}\approx{\frac{\mathcal{L}\circ\mathbf{f}\circ\mathbf{r}\!\left(\mathbf{X}+\delta\cdot\mathbf{e}_d\right)-\mathcal{L}\circ\mathbf{f}\circ\mathbf{r}\!\left(\mathbf{X}-\delta\cdot\mathbf{e}_d\right)}{2\times\delta}}$, where $\mathbf{e}_d$ is a $D$-dimensional vector with the $d$-th dimension set to be $1$ and all the others to be $0$. In general, every step of such update may randomly select a subset of all $D$ dimensions for efficiency considerations, so our optimization algorithm is a form of stochastic coordinate descent. This is reminiscent of~\cite{Chen_2017_ZOO}, where each step updates the values of a random subset of pixel values. Also following~\cite{Chen_2017_ZOO}, we use the Adam optimizer~\cite{Kingma_2015_Adam} instead of standard gradient descent for its faster convergence.

\newcommand{\colwidthA}{1.0cm}
\newcommand{\colwidthB}{0.7cm}
\begin{table*}[!btp]
\centering
\begin{tabular}{|l||R{\colwidthA}|R{\colwidthB}||R{\colwidthA}|R{\colwidthB}
                   |R{\colwidthA}|R{\colwidthB}|R{\colwidthA}|R{\colwidthB}||R{\colwidthA}|R{\colwidthB}|}
\hline
{Attacking}     & \multicolumn{2}{c||}{Image}           & \multicolumn{2}{c|}{Surface N.}
                & \multicolumn{2}{c|}{Illumination}     & \multicolumn{2}{c||}{Material}
                & \multicolumn{2}{c|}{Combined}         \\
\cline{2-11}
{Perturbations} & Succ.    & $p$    & Succ.    & $p$    & Succ.    & $p$    & Succ.    & $p$    & Succ.    & $p$    \\
\hline\hline
On AlexNet & $100.00$ & $5.7$ & $89.27$ & $10.8$ & $29.61$ & $25.8$ & $18.88$ & $25.8$ & $94.42$ & $18.1$ \\
\hline
On ResNet-34 & $99.57$ & $5.1$ & $88.41$ & $9.3$ & $14.16$ & $29.3$ & $ 3.43$ & $55.2$ & $94.85$ & $16.4$ \\
\hline
\end{tabular}
\smallskip
\caption{
    Effect of white-box adversarial attacks on ShapeNet object classification. By {\em combined}, we allow the three sets of physical parameters to be perturbed jointly. {\bf Succ.} denotes the success rate of attacks ($\%$, higher is better), and $p$ is the perceptibility value (unit: $10^{-3}$, lower is better). All $p$ values are measured in the image space, {\em i.e.}, they are directly comparable.
}
\label{Tab:Classification}
\end{table*}

\subsection{Perceptibility}

The goal of an adversarial attack is to produce a visually imperceptible perturbation, so that the network makes incorrect predictions after it is added to the original image. Given a rendering model ${\mathbf{Y}}={\mathbf{r}\!\left(\mathbf{X}\right)}$ and an added perturbation $\Delta\mathbf{X}$, the perturbation added to the rendered image is: ${\Delta\mathbf{Y}}={\mathbf{r}\!\left(\mathbf{X}+\Delta\mathbf{X}\right)-\mathbf{r}\!\left(\mathbf{X}\right)}$.

There are in general two ways of computing perceptibility. One of them works directly on the rendered image, which is similar to the definition in~\cite{Szegedy_2014_Intriguing}\cite{Moosavi_2015_DeepFool}: ${p}\doteq{p\!\left(\Delta\mathbf{Y}\right)}={\left(\frac{1}{W_\mathrm{N}\times H_\mathrm{N}}{\sum_{w=1}^{W_\mathrm{N}}}{\sum_{h=1}^{H_\mathrm{N}}}\left\|\Delta\mathbf{y}_{w,h}\right\|_2^2\right)^{1/2}}$, where $\mathbf{y}_{w,h}$ is a $3$-dimensional vector representing the RGB intensities (normalized in $\left[0,1\right]$) of a pixel. Similarly, we can also define the perceptibility values for each set of physical parameters, {\em e.g.}, ${p\!\left(\Delta\mathbf{N}\right)}={\left(\frac{1}{W_\mathrm{N}\times H_\mathrm{N}}{\sum_{w=1}^{W_\mathrm{N}}}{\sum_{h=1}^{H_\mathrm{N}}}\left\|\Delta\mathbf{n}_{w,h}\right\|_2^2\right)^{1/2}}$.

We take $p\!\left(\Delta\mathbf{Y}\right)$ as the major criterion of {\em visual} imperceptibility. Because of continuity, this can guarantee that all physical perturbations are sufficiently small as well.
An advantage of placing the perceptibility constraint on pixels is that it allows a fair comparison of the attack success rates between image space attacks and physical space attacks. It also allows a direct comparison between attacks on different physical parameters.
One potential disadvantage of placing the perceptibility constraint on physical parameters is that different physical parameters have different units and ranges. 
For example, the value range of RGB is $[0, 255]$, whereas that of spatial translation is $(-\infty, \infty)$. 
It is not directly obvious how to find a common threshold for different physical parameters. 

When using the differentiable renderer, in order to guarantee imperceptibility, we constrain the RGB intensity changes on the image layer. In each iteration, after a new set of physical perturbations are generated, we check all pixels on the re-rendered image, and any perturbations exceeding a fixed threshold $U=18$ from the original image is {\em truncated}. Truncations cause the inconsistency between the physical parameters and the rendered image and risk failures in attacking. To avoid frequent truncations, we set the learning rate $\eta$ to be small, which consequently increases the number of iterations needed to attack the network.

When using the non-differentiable renderer, we pursue an alternative approach by adding another term $\left\|\Delta\mathbf{Y}\right\|_2^2$ into the loss function (weighted by $\lambda$)~\cite{Chen_2017_ZOO,Carlini_2017_Towards}, such that optimization can balance between attack success and perceptibility.

\subsection{Interpreting Image Space Adversaries in Physical Space}

We do a reality check to confirm that image-space adversaries are almost never consistent with the non-local physical perturbations according to our (admittedly imperfect) rendering model. They are, of course, consistent with per-pixel changes of albedo. 

We first find a perturbation $\Delta\mathbf{Y}$ in the image space, and then compute a perturbation in the physical space, $\Delta\mathbf{X}$, that corresponds to $\Delta\mathbf{Y}$. This is to set the optimization goal in the image space instead of the output space, though the optimization process is barely changed. Note that we are indeed pursuing interpreting $\Delta\mathbf{Y}$ in the physical space. Not surprisingly, as we will show in experiments, the reconstruction loss $\left\|\mathbf{Y}+\Delta\mathbf{Y}-\mathbf{r}\!\left(\mathbf{X}+\Delta\mathbf{X}\right)\right\|_1$ does not go down, suggesting that approximations of $\Delta\mathbf{Y}$ in the physical space either do not exist, or cannot be found by the currently available optimization methods such as FGSM.

\section{Experiments}
\label{Experiments}

\subsection{3D Object Classification}
\label{Experiments:Classification}

\begin{figure}[!t]
\begin{center}
    \includegraphics[width=\linewidth]{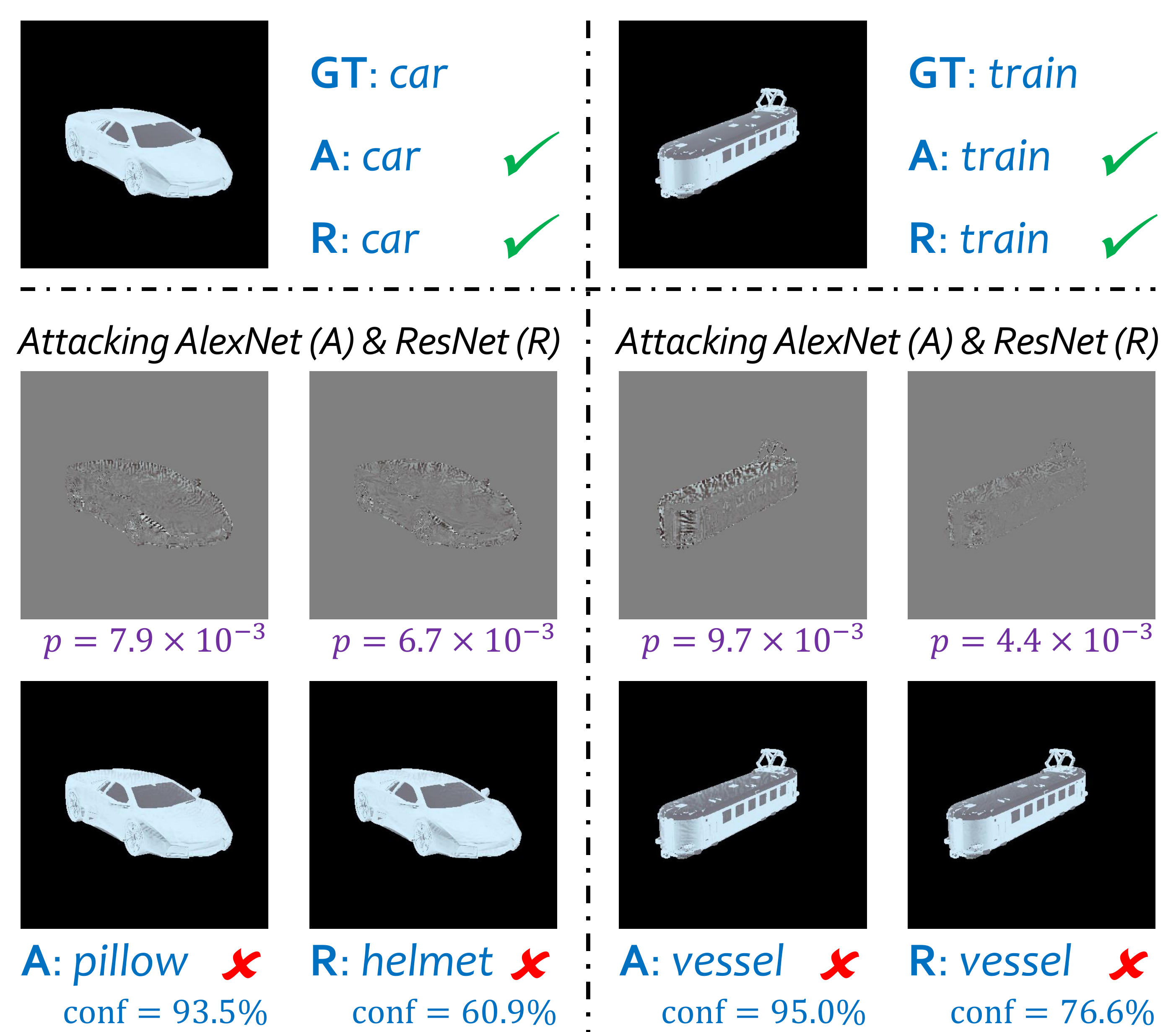}
\end{center}
\caption{
    Examples of physical-space adversaries in 3D object classification on ShapeNet (using a differentiable renderer). In each example, the top row shows the original testing image, which is correctly classified by both AlexNet (A) and ResNet (R). The following two rows show the perturbations and the attacked image, respectively. All perturbations are magnified by a factor of $\mathbf{5}$ and shifted by $\mathbf{128}$. $p$ is the perceptibility value, and $\mathrm{conf}$ is the confidence (post-softmax output) of the prediction.
}
\label{Fig:ClassificationExamples}
\end{figure}

\begin{figure*}[!t]
\centering
    \includegraphics[width=\textwidth]{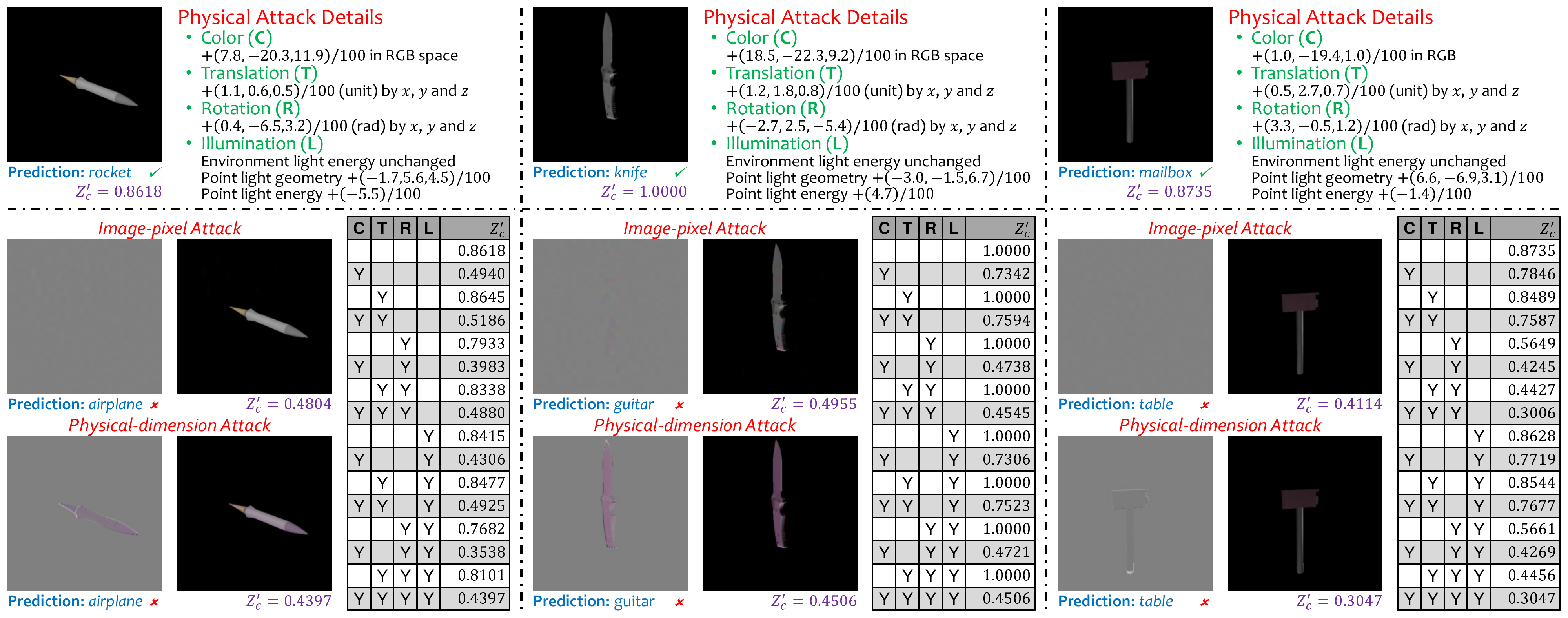}
\caption{
    Examples of image-space and physical-space adversaries in 3D object classification on ShapeNet (using a non-differentiable renderer). In each example, the top row contains the original testing image and the detailed description of mid-level physical operations that can cause classification to fail. In the bottom row, we show the perturbations and attacked images in both attacks. $Z_c'$ is the confidence (post-softmax output) of the true class. For each case, we also show results with different combinations of physical attacks in a table (a \textsf{Y} indicates the corresponding attack is on).
}
\label{Fig:ClassificationExamples2}
\end{figure*}

3D object recognition experiments are conducted on the ShapeNetCore-v2 dataset~\cite{Chang_2015_ShapeNet}, which contains $55$ rigid object categories, each with various 3D models. Two popular deep neural networks are used: an $8$-layer AlexNet~\cite{Krizhevsky_2012_ImageNet} and a $34$-layer deep residual network~\cite{He_2016_Deep}. Both networks are pre-trained on the ILSVRC2012 dataset~\cite{Russakovsky_2015_ImageNet}, and fine-tuned in our training set for $40$ epochs using batch size 256. The learning rate is $0.001$ for AlexNet and $0.005$ for ResNet-34.

We experiment with both a differentiable renderer~\cite{Liu_2017_Material} and a non-differentiable renderer ~\cite{Blender_2017_Blender}, and as a result there are some small differences in the experimental setup, despite the shared settings described above.

For the {\bf differentiable renderer}, we randomly sample $125$ 3D models from each class, and select $4$ fixed viewpoints for each object, so that each category has $500$ training images. Similarly, another randomly chosen $50\times4$ images for each class are used for testing. AlexNet and ResNet-34 achieve $73.59\%$ and $79.35\%$ top-$1$ classification accuracies, respectively. These numbers are comparable to the single-view baseline accuracy reported in~\cite{Su_2015_Multi}. For each class, from the correctly classified testing samples, we choose $5$ images with the highest classification probabilities on ResNet-34, and filter out $22$ of them which are incorrectly classified by AlexNet, resulting in a target set of $233$ images. The attack algorithm is the iterative version of FGSM~\cite{Goodfellow_2015_Explaining}. We use the SGD optimizer with momentum $0.9$ and weight decay $10^{-4}$, and the maximal number of iterations is $120$. Learning rate is $0.002$ for attacking image space, $0.003$ for attacking illumination and material, and $0.004$ for attacking surface normal.

For the {\bf non-differentiable renderer}, we render images with an azimuth angle uniformly sampled from $\left[0,\pi\right)$, a fixed elevation angle of $\pi/9$ and a fixed distance of $1.8$. AlexNet gives a $65.89\%$ top-$1$ testing set classification accuracy, and ResNet-34 achieves an even higher number of $68.88\%$. Among $55$ classes, we find $51$ with at least two images correctly classified. From each of them, we choose the two correct testing cases with the highest confidence score and thus compose a target set with $102$ images. The attack algorithm is ZOO~\cite{Chen_2017_ZOO} with ${\delta}={10^{-4}}$, ${\eta}={2\times10^{-3}}$ and ${\lambda}={0.1}$. The maximal number of iterations is $500$ for AlexNet and $200$ for ResNet-34.

\subsubsection{Differentiable Renderer Results}

\noindent
First, we demonstrate in Table~\ref{Tab:Classification} that adversaries widely exist in the image space -- as researchers have explored before~\cite{Szegedy_2014_Intriguing}\cite{Moosavi_2015_DeepFool}, it is easy to confuse the network with small perturbations. In our case, the success rate is at or close to $100\%$ and the perceptibility does not exceed $10^{-2}$.

The next study is to find the correspondence of these image-space perturbations in the physical space. We tried the combination of 3 learning rates ($10^{-3}, 10^{-4}, 10^{-5}$) and 2 optimizers (SGD, Adam). However, for AlexNet, the objective ($\ell_1$-distance) remains mostly constant; the malicious label after image-space attack is kept in only $8$ cases, and in the vast majority cases, the original true label of the object is recovered. Therefore, using the current optimization method and rendering model, it is very difficult to find physical parameters that are approximately rendered into these image-space adversaries. This is expected, as physical parameters often have a non-local effect on the image.

Finally we turn to directly generating adversaries in the physical space. As shown in Table~\ref{Tab:Classification}, this is much more difficult than in the image space -- the success rate becomes lower and large perceptibility values are often observed on the successful cases. Typical adversarial examples generated in the physical space are shown in Figure~\ref{Fig:ClassificationExamples}. Allowing all physical parameters to be jointly optimized ({\em i.e.}, the {\em combined} strategy) produces the highest success rate. Among the three sets of physical parameters, attacking surface normals is more effective than the other two. This is expected, as using local perturbations is often easier in attacking deep neural networks~\cite{Goodfellow_2015_Explaining}. The surface normal matrix shares the same dimensionality with the image lattice, and changing an element in the matrix only has very local impact on the rendered image. In comparison, illumination and material are both global properties of the 3D scene or the object, so tuning each parameter will cause a number of pixels to be modified, hence less effective in adversarial attacks. 

We also examined truncation during the attack. For ResNet-34, on average, only $6.3$, $1.6$, $0$ pixels were ever truncated for normal, illumination, material throughout the $120$ iterations of attack. This number of truncation is relatively small comparing to the size of the rendered image ($448\times 448$). Therefore, the truncation is unlikely to contribute much to the attack. 

\subsubsection{Non-differentiable Renderer Results}

\noindent
We first report quantitative results with two settings, {\em i.e.}, attacking the image space and the physical space. Similarly, image-space adversaries are relatively easy to find. Among all $102$ cases, $99$ of them are successfully attacked within $500$ steps on AlexNet, and all of them within $200$ steps on ResNet-34. On the other hand, physical-space adversaries are much more difficult to construct. Using the same numbers of steps ($500$ on AlexNet and $200$ on ResNet-34), the numbers of success attacks are merely $14$ and $6$ respectively.

We show several successful cases of image-space and physical-space attacks in Figure~\ref{Fig:ClassificationExamples2}. One can see quite different perturbation patterns from these two scenarios. An image-space perturbation is the sum of pixel-level differences, {\em e.g.}, even the intensities of two adjacent pixels can be modified individually, thus it is unclear if these images can really appear in the real world, nor can we diagnose the reason of failure. On the other hand, a physical-space perturbation is generated using a few mid-level operations such as slight rotation, translation and minor lighting changes. In theory, these adversaries can be instantiated in the physical world using a fine-level robotic controlling system.

Another benefit of generating physical-dimension adversaries lies in the ability of diagnosing vision algorithms. We use the cases shown in Figure~\ref{Fig:ClassificationExamples2} as examples. There are $14$ changeable physical parameters, and we partition them into $4$ groups, {\em i.e.}, the environment illumination ($5$ parameters), object rotation, position and color ($3$ parameters each). We enumerate all $2^4$ subsets of these parameters, and thus generate $2^4$ perturbations by only applying the perturbations in the subsets. It is interesting to see that in the first case, the effects of different perturbations are almost additive, {\em e.g.}, the joint attack on color and rotation has roughly the same effect as the sum of individual attacks. However, this is not always guaranteed. In the second case, for example, we find that attacking rotation alone produces little effect, but adding it to color attack causes a dramatic accuracy drop of $26\%$. On the other hand, the second case is especially sensitive to color, and the third one to rotation, suggesting that different images are susceptible to attacks in different subspaces. It is the interpretability of the physical-dimension attacks that provides the possibility to diagnose these cases at a finer level.

\renewcommand{\colwidthA}{1.0cm}
\renewcommand{\colwidthB}{0.7cm}
\begin{table*}[t]
\centering
\begin{tabular}{|l||R{\colwidthA}|R{\colwidthB}||R{\colwidthA}|R{\colwidthB}
                   |R{\colwidthA}|R{\colwidthB}|R{\colwidthA}|R{\colwidthB}||R{\colwidthA}|R{\colwidthB}|}
\hline
{Attacking}     & \multicolumn{2}{c||}{Image}           & \multicolumn{2}{c|}{Surface N.}
                & \multicolumn{2}{c|}{Illumination}     & \multicolumn{2}{c||}{Material}
                & \multicolumn{2}{c|}{Combined}         \\
\cline{2-11}
{Perturbations} & Succ.    & $p$    & Succ.    & $p$    & Succ.    & $p$    & Succ.    & $p$    & Succ.    & $p$    \\
\hline\hline
On IEP~\cite{Johnson_2017_Inferring} & $96.33$ & $2.1$ & $83.67$ & $6.8$ & $48.67$ & $9.5$ & $8.33$ & $12.3$ & $90.67$ & $8.8$ \\
\hline
\end{tabular}
\smallskip
\caption{
    Effect of white-box adversarial attacks on CLEVR visual question answering. By {\em combined}, we allow the three sets of physical parameters to be perturbed jointly. {\bf Succ.} denotes the success rate of attacks ($\%$, higher is better) of giving a correct answer, and $p$ is the perceptibility value (unit: $10^{-3}$, lower is better). All $p$ values are measured in the image space, {\em i.e.}, they are directly comparable.
}
\label{Tab:Answering}
\end{table*}

\subsection{Visual Question Answering}
\label{Experiments:Answering}

We extend our experiments to a more challenging vision task -- visual question answering. Experiments are performed on the recently released CLEVR dataset~\cite{Johnson_2017_CLEVR}. This is an engine that can generate an arbitrary number of 3D scenes with meta-information (object configuration). Each scene is also equipped with multiple generated questions, {\em e.g.}, asking for the number of specified objects in the scene, or if the object has a specified property.

The baseline algorithm is named Inferring and Executing Programs (IEP)~\cite{Johnson_2017_Inferring}. It applies an LSTM to parse each question into a tree-structure program, which is then converted into a neural module network~\cite{Andreas_2016_Neural} that queries the visual features.
We use the released model without training it by ourselves.
We randomly pick up $100$ testing images, on which all associated questions are correctly answered, as the target images.

The settings for generating adversarial perturbations are the same as in the object classification experiments: when using the differentiable renderer, the iterative FGSM is used, and three sets of physical parameters are attacked either individually or jointly; when using the non-differentiable renderer, the ZOO algorithm~\cite{Chen_2017_ZOO} is used with ${\delta}={10^{-3}}$, ${\eta}={10^{-2}}$, ${\lambda}={0.5}$.

\begin{figure}[t]
\begin{center}
    \includegraphics[width=\linewidth]{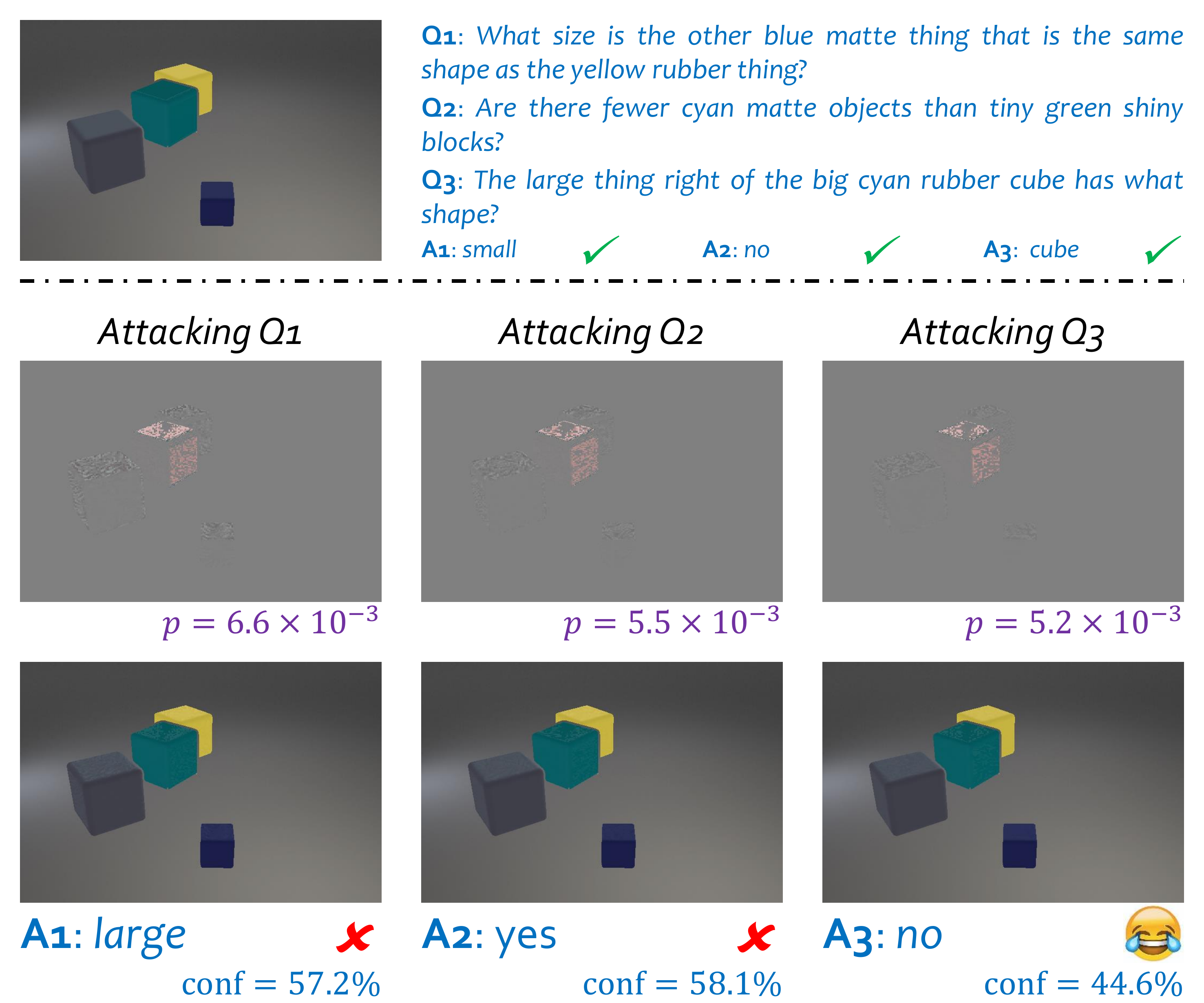}
\end{center}
\caption{
    An example of physical-space adversaries in 3D visual question answering on CLEVR (using a differentiable renderer). In each example, the top row shows a testing image and three questions, all of which are correctly answered. The following two rows show the perturbations and the attacked image, respectively. All perturbations are magnified by a factor of $\mathbf{5}$ and shifted by $\mathbf{128}$. $p$ is the perceptibility value, and $\mathrm{conf}$ is the confidence (post-softmax output) of choosing this answer.
}
\label{Fig:AnsweringExamples}
\end{figure}

\subsubsection{Differentiable Renderer Results}

\noindent
Results are shown in Table~\ref{Tab:Answering}. We observe similar phenomena as in the classification experiments. This is expected, since after the question is parsed and a neural module network is generated, attacking either the image or the physical space is essentially equivalent to that in the classification task. Some typical examples are shown in Figure~\ref{Fig:AnsweringExamples}.

A side note comes from perturbing the material parameters. Although some visual questions are asking about the material ({\em e.g.}, {\em metal} or {\em rubber}) of an object, the success rate of this type of questions does not differ from that in attacking other questions significantly. This is because we are constraining perceptibility, which does not allow the material parameters to be modified by a large value.

A significant difference of visual question answering comes from the so called {\em language prior}. With a language parser, the network is able to clinch a small subset of answers without looking at the image, {\em e.g.}, when asked about the {\em color} of an object, it is very unlikely for the network to answer {\em yes} or {\em three}. Yet we find that sometimes the network can make such ridiculous errors. For instance, in the rightmost column of Figure~\ref{Fig:AnsweringExamples}, when asked about the {\em shape} of an object, the network answers {\em no} after a {\em non-targeted} attack.

\subsubsection{Non-differentiable Renderer Results}

\begin{figure}[!btp]
\centering
    \includegraphics[width=\linewidth]{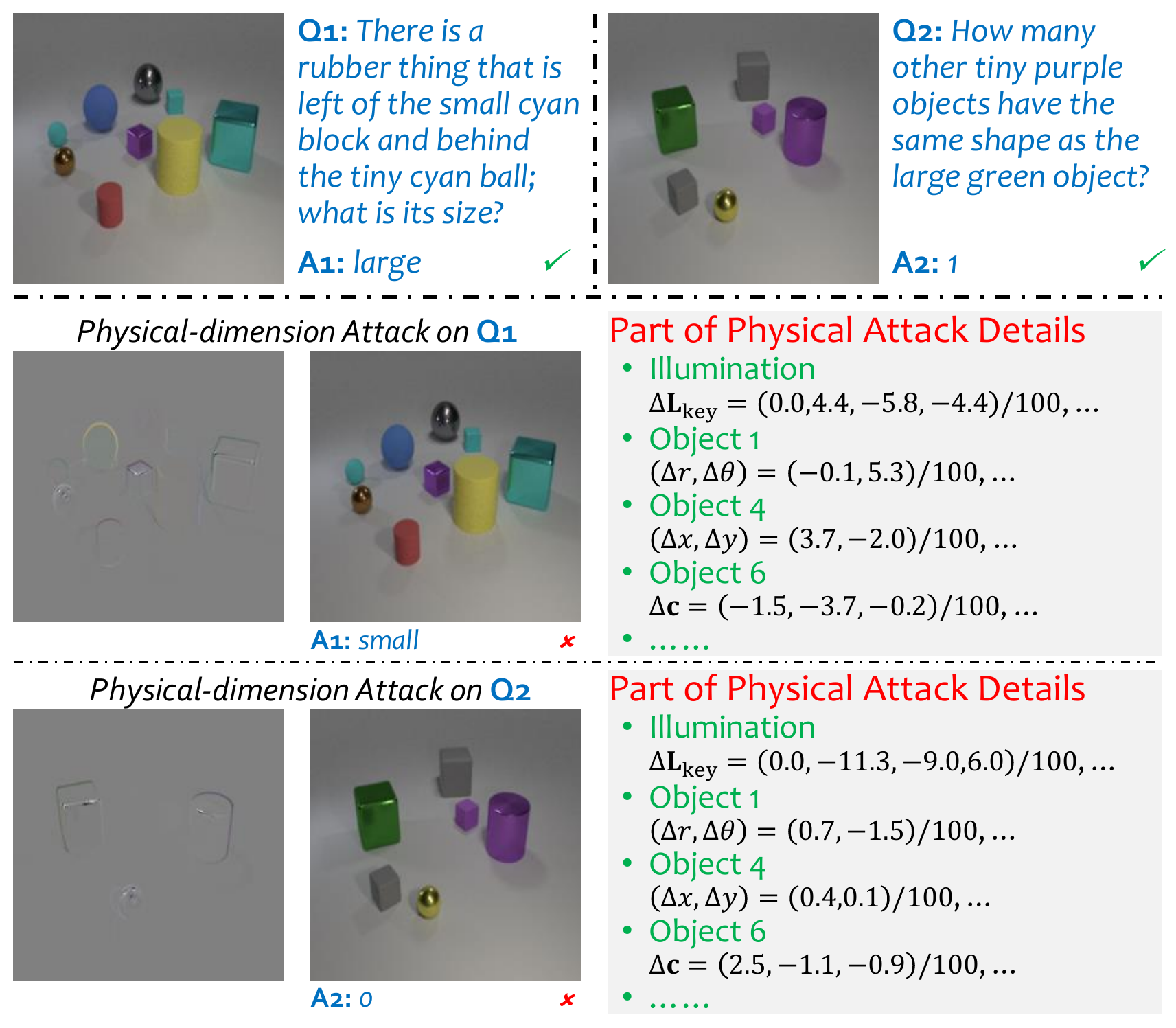}
\caption{
    Examples of physical-space adversaries in 3D visual question answering on CLEVR (using a non-differentiable renderer). In each example, the top row contains a testing image and three questions. In the bottom row, we show the perturbations and attacked images. Detailed description of physical attacks on selective dimensions are also provided. All units of physical parameters follow the default setting in Blender.
}
\label{Fig:AnsweringExamples2}
\end{figure}

\noindent
We observe quite similar results as in ShapeNet experiments. It is relatively easy to find image-space adversaries, as our baseline successfully attacks $66$ out of $100$ targets within $500$ steps, and $93$ within $1\rm{,}200$ steps. Due to computational considerations, we set $500$ to be the maximal step in our attack experiment, but only find $22$ physical-space adversaries. This is expected, since visual question answering becomes quite similar to classification after the question is fixed.

We show two successfully attacked examples in Figure~\ref{Fig:AnsweringExamples2}. Unlike ShapeNet experiments, color plays an important role in CLEVR, as many questions are related to filtering/counting objects with specified colors. We find that in many cases, our algorithm achieves success by mainly attacking the color of the key object ({\em i.e.} that asked in the question). This could seem problematic, as generated adversaries may threaten the original correct answer. But according to our inspection, the relatively big $\lambda$ we chose ensured otherwise. Nevertheless, this observation is interesting because our algorithm does not know the question ({\em i.e.}, IEP is a black-box) or the answer ({\em i.e.}, each answer is simply a class ID), but it automatically tries to attack the weakness ({\em e.g.}, color) of the vision system.

\section{Conclusions}
\label{Conclusions}

In this paper, we generalize adversarial examples beyond the 2D image pixel intensities to 3D physical parameters. We are mainly interested to know: are neural networks vulnerable to perturbation on these intrinsic parameters that define a 3D scene, just like they are vulnerable to artificial noise added to the image pixels?

To study this, we plug a rendering module in front of the state-of-the-art deep networks, in order to connect the underlying 3D scene with the perceived 2D image. We are then able to conduct gradient based attacks on this more complete vision pipeline. Extensive experiments in object classification and visual question answering show that directly constructing adversaries in the physical space is effective, but the success rate is lower than that in the image space, and much heavier perturbations are required for successful attacks. To the best of our knowledge, ours is the first work to study imperceptible adversarial examples in 3D, where each dimension of the adversarial perturbation has clear meaning in the physical world.

Going forward, we see three potential directions for further research.
First, as a side benefit, our study may provide practical tools to diagnose vision algorithms, especially evaluating the robustness in some interpretable dimensions such as color, lighting and object movements.
Second, in 3D vision scenarios, we show the promise to defend the deep neural networks against 2D adversaries by interpreting an image in the physical space, so that the adversarial effects are weakened or removed after re-rendering.
Third, while our pipeline will continue to benefit from higher quality rendering, we also acknowledge the necessity to test out our findings in real-world scenarios.

\ifcvprfinal
\vspace{-0.5cm}
\paragraph{Acknowledgments}
We thank Guilin Liu, Cihang Xie, Zhishuai Zhang and Yi Zhang for discussions.
This research is supported by IARPA D17PC00342 and a gift from YiTu.
\fi

{\small
\bibliographystyle{ieee}
\bibliography{egbib}
}

\input{supp}

\end{document}

%% file: supp.tex
\clearpage
\onecolumn
\begin{center}
    {\Large \bf Supplementary Material \par}
    \vspace*{24pt}
\end{center}
\appendix

\section{Attack Curves with Different (Differentiable or Non-Differentiable) Renderers}

\begin{figure}[H]
\centering
\includegraphics[width=0.33\linewidth]{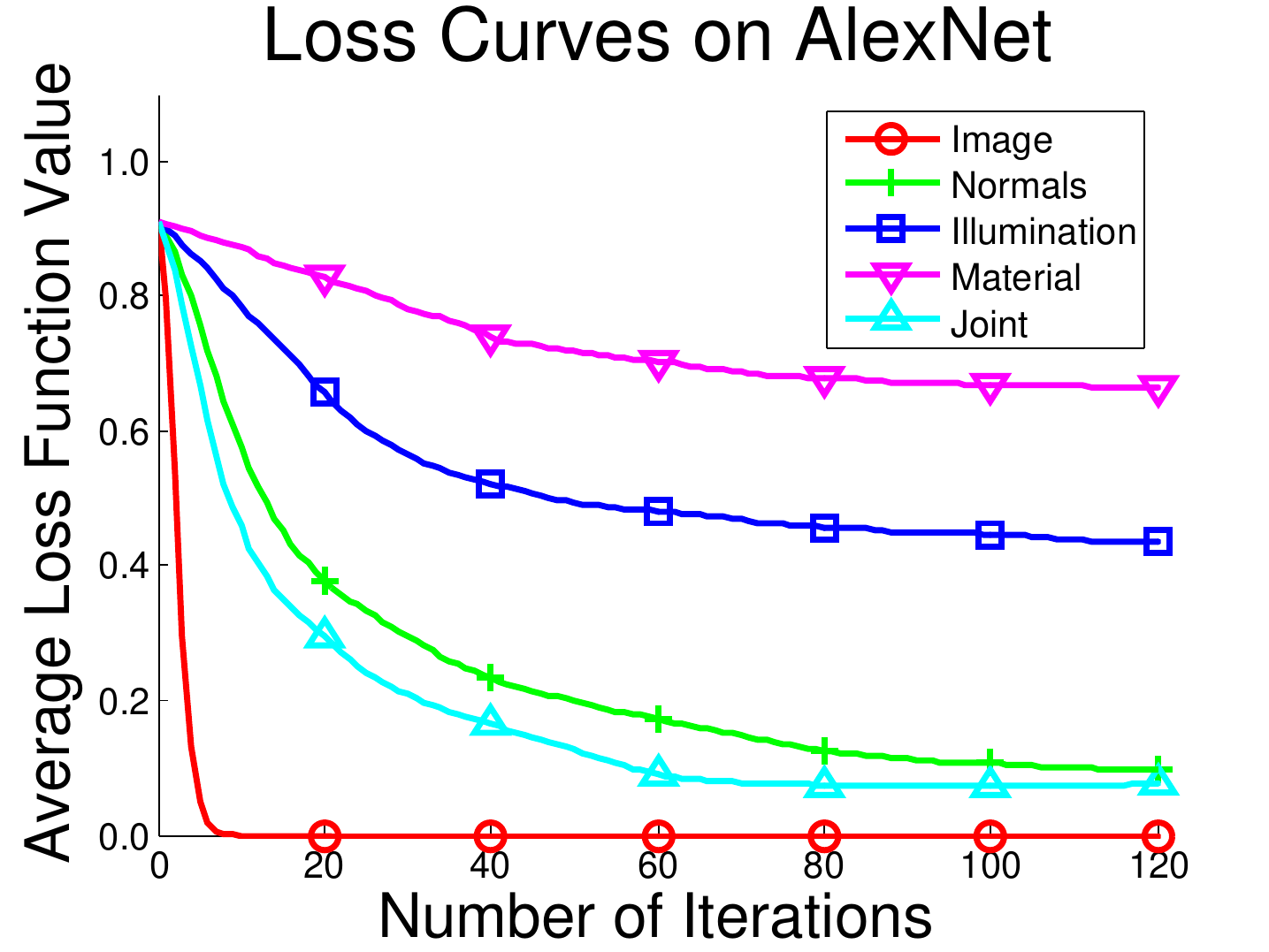}
\includegraphics[width=0.33\linewidth]{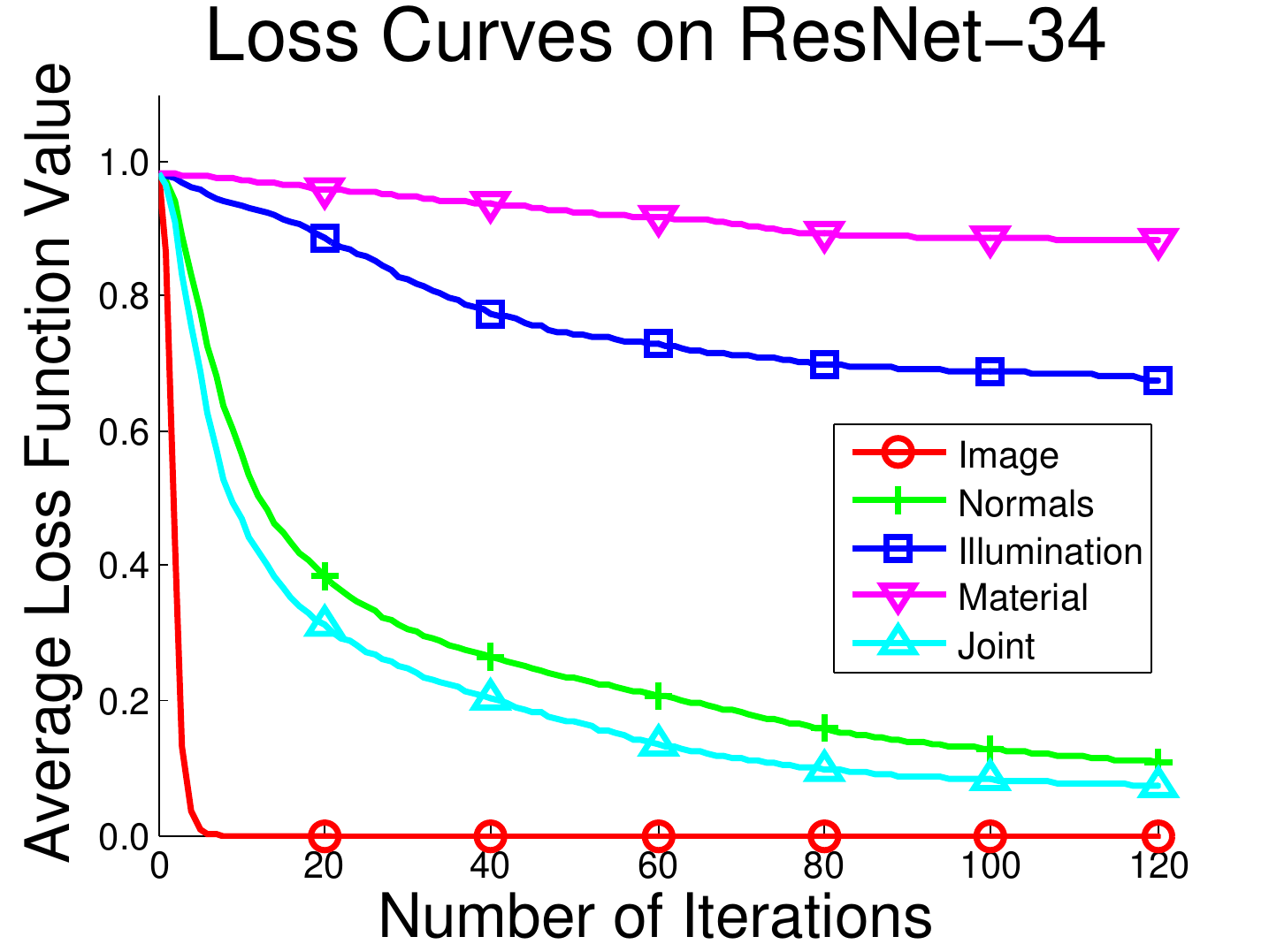}
\caption{Attack curves for 3D object classification with a differentiable renderer.}
\label{fig:diff}
\end{figure}

In Figure~\ref{fig:diff}, we plot how the average loss function value (probability of the original class, after softmax) changes with respect to the number of attack iterations. Image-space attacks often succeed very quickly, whereas physical-space attacks are much slower yet more difficult, especially for the factors of illumination and material.

\begin{figure}[H]
\centering
\includegraphics[width=0.33\linewidth]{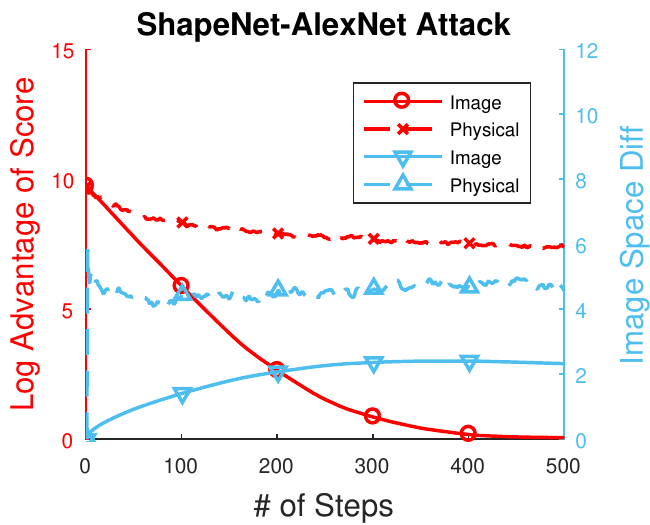}
\includegraphics[width=0.33\linewidth]{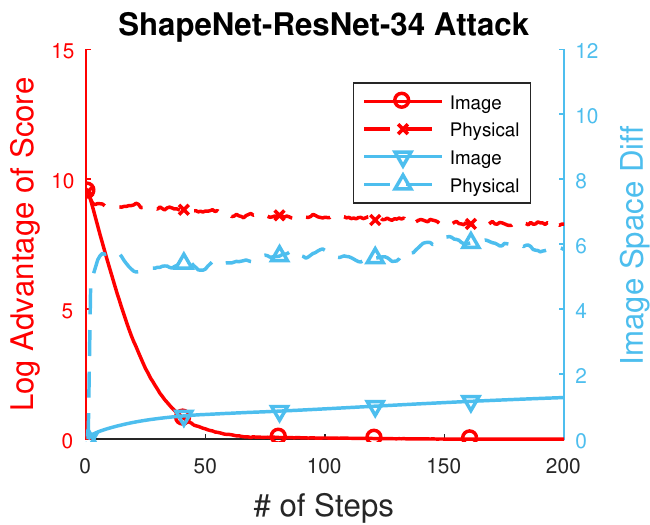}
\includegraphics[width=0.33\linewidth]{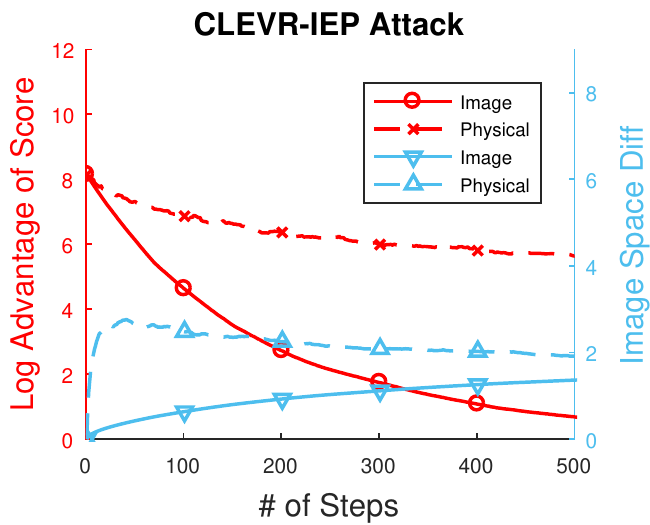}
\caption{Attack curves for 3D object classification and visual question answering with a non-differentiable renderer.}
\label{fig:nondiff}
\end{figure}

In Figure~\ref{fig:nondiff}, we plot how the average log-probability advantage (red) and image-space Euclidean distance (blue) change with respect to the number of attack iterations. An average log-probability advantage of $0$ means that all images have been attacked successfully. Physical-space attacks are much more difficult to succeed and also require a much larger perceptibility.

From these curves, we can conclude that physical-space attacks especially adding factors with clear physical meanings are much more difficult. This is arguably because most of these attacks impact the values of more than one pixels in the image space, which raises higher difficulties to the optimizers ({\em e.g.}, gradient-descent-based). We should also note that, with a more powerful optimizer, it is possible to find more adversarial examples in the physical world.